\documentclass[10pt,twocolumn,letterpaper]{article}

\usepackage{cvpr}
\usepackage{times}
\usepackage{epsfig}
\usepackage{graphicx}
\usepackage{amsmath}
\usepackage{amssymb}

\usepackage{float}
\usepackage{makecell}
\usepackage[breaklinks=true,bookmarks=false]{hyperref}
\cvprfinalcopy

\def\cvprPaperID{1728}

\begin{document}
\title{Learning Oracle Attention for High-fidelity Face Completion}

\author{Tong Zhou$^1$ \quad Changxing Ding$^{1}$
 \quad Shaowen Lin$^1$ \quad Xinchao Wang$^2$ \quad Dacheng Tao$^3$\\
$^1$ South China University of Technology \quad $^2$ Stevens Institute of Technology \\
$^3$ UBTECH Sydney AI Centre, School of Computer Science, Faculty of Engineering,\\
The University of Sydney, Darlington, NSW 2008, Australia\\
{\tt\small 201821011282@mail.scut.edu.cn \quad chxding@scut.edu.cn \quad } \\
{\tt\small eeswlin@mail.scut.edu.cn \quad xinchao.w@gmail.com  \quad dacheng.tao@sydney.edu.au}
}

\maketitle
\thispagestyle{empty}

\begin{abstract}
   High-fidelity face completion is a challenging task due to the rich and subtle facial textures involved. What makes it more complicated is the correlations between different facial components, for example, the symmetry in texture and structure between both eyes. While recent works adopted the attention mechanism to learn the contextual relations among elements of the face, they have largely overlooked the disastrous impacts of inaccurate attention scores; in addition, they fail to pay sufficient attention to key facial components, the completion results of which largely determine the authenticity of a face image. Accordingly, in this paper, we design a comprehensive framework for face completion based on the U-Net structure. Specifically, we propose a dual spatial attention module to efficiently learn the correlations between facial textures at multiple scales; moreover, we provide an oracle supervision signal to the attention module to ensure that the obtained attention scores are reasonable. Furthermore, we take the location of the facial components as prior knowledge and impose a multi-discriminator on these regions, with which the fidelity of facial components is significantly promoted. Extensive experiments on two high-resolution face datasets including CelebA-HQ and Flickr-Faces-HQ demonstrate that the proposed approach outperforms state-of-the-art methods by large margins.
\end{abstract}
\section{Introduction}
\begin{figure}
\begin{center}
\includegraphics[width=0.8\linewidth]{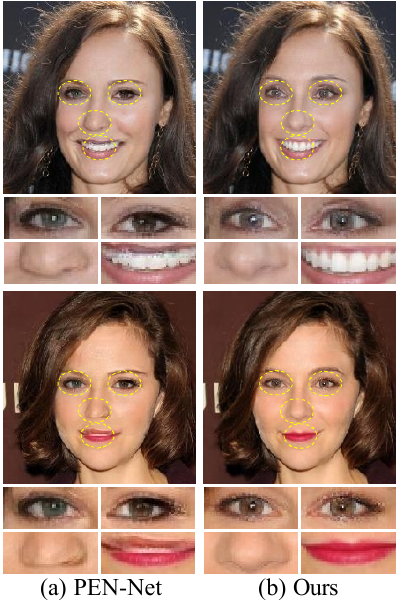}
\end{center}
   \caption{Face completion results for images ($256 \times 256$) with center mask ($128 \times 128$). In each row from left to right: (a) the result taken from the paper of PEN-Net~\cite{zeng2019learning}. By zooming in, we can observe the color discrepancy between two eyes, along with distortions in the nose and mouth regions. (b) The results of our method. It can be seen that our method indeed produces face images with high fidelity.}
\label{fig:first}
\end{figure}
Image inpainting refers to filling in the missing pixels in an image with the expectation that the recovered image will be visually realistic. This process not only requires that the filled textures themselves be meaningful, but also seeks semantic consistency between the filled area and the context. Image inpainting is widely applied to photo restoration, image editing, and object removal, among many other tasks.

Face completion, as a branch of image inpainting, focuses on filling the missing regions of a human face, and turns out to be a challenging task. The reasons behind lie in two aspects. First, the human face contains rich and subtle textures that also differ dramatically across persons, meaning that it is difficult to perfectly restore these diverse facial textures. Second, there are close correlations between facial components, making the fidelity of the image more vulnerable to the semantic consistency between the facial components.
Take one very recent work~\cite{zeng2019learning} for example, its generated images are satisfactory in facial structure, but still suffer from small artifacts in facial components and semantic inconsistency.
As illustrated in Figure~\ref{fig:first}, the two eyes in the same face are different in color; moreover, small distortions in the nose and mouth areas can also be observed. These flaws have a substantial impact on the authenticity of the overall visual effect.

Recently, convolutional neural network (CNN) based methods have become the mainstream approach to image inpainting~\cite{yang2017high,yan2018shift,li2019boosted, wang2018image,zhang2018semantic}. In order to generate visually realistic images, existing approaches typically adopt the non-local schemes that make use of contextual relations to fill in the missing pixels~\cite{yu2018generative, sagong2019pepsi, zheng2019pluralistic, Song2018ECCV}. However, due to the lack of direct supervision on the obtained attention scores, the learned relations are insufficiently reliable, meaning that these methods may generate distorted textures.

Moreover, different types of structural information have been extracted by networks in order to act as prior knowledge for assisting image inpainting; for example, segmentation maps~\cite{song2018spg}, object contours~\cite{xiong2019foreground}, edge~\cite{nazeri2019edgeconnect}, facial landmarks~\cite{song2018geometry}, and face parsing~\cite{li2017generative}. While these methods focus on the correctness of the structural information, they also ignore the quality of the textures on key areas in the image (e.g. facial components in face images).

In this paper, we propose a comprehensive framework to handle the above issues. Inspired by recent progress in attention models~\cite{zhang2019self, fu2019dual}, we propose a Dual Spatial Attention (DSA) model that comprises foreground self-attention and foreground-background cross-attention modules. To capture contextual information across different scales, we apply it to multiple layers in the network. Compared with attention models introduced in recent works~\cite{yu2018generative, liu2019coherent}, DSA has two key advantages: first, it is more efficient and can capture more comprehensive contextual information; second, we impose an oracle supervision signal to ensure that the attention scores produced by DSA are reasonable. With the help of DSA, our approach obtains semantically consistent face completion results, as illustrated in Figure~\ref{fig:first}.

Furthermore, we also extract facial landmarks from the ground truth image to act as prior knowledge. Rather than imposing constraints to ensure that the facial landmarks between the recovered and ground-truth images are consistent, we use facial landmarks to locate four key facial components: i.e., both eyes, the nose, and the mouth. Subsequently, we train four discriminators for each recovered facial component. By benefiting from adversarial learning on the specified locations, our generator pays more attention to the textures of each key facial component. As a result, our proposed approach can generate more visually realistic textures, as shown in Figure~\ref{fig:first}. Since all discriminators are removed during testing, our approach does not yield any drop in efficiency.

We conduct a number of experiments on high-resolution face datasets, including CelebA-HQ~\cite{Karras2017Progressive} and Flickr-Faces-HQ~\cite{karras2019style}. Quantitative and qualitative comparisons demonstrate that our proposed approach outperforms state-of-the-art methods by large margins.
\section{Related Work}
\pagestyle{empty}
\begin{figure*}
\begin{center}
%\fbox{\rule{0pt}{2in} \rule{.9\linewidth}{0pt}}
\includegraphics[width=.9\linewidth]{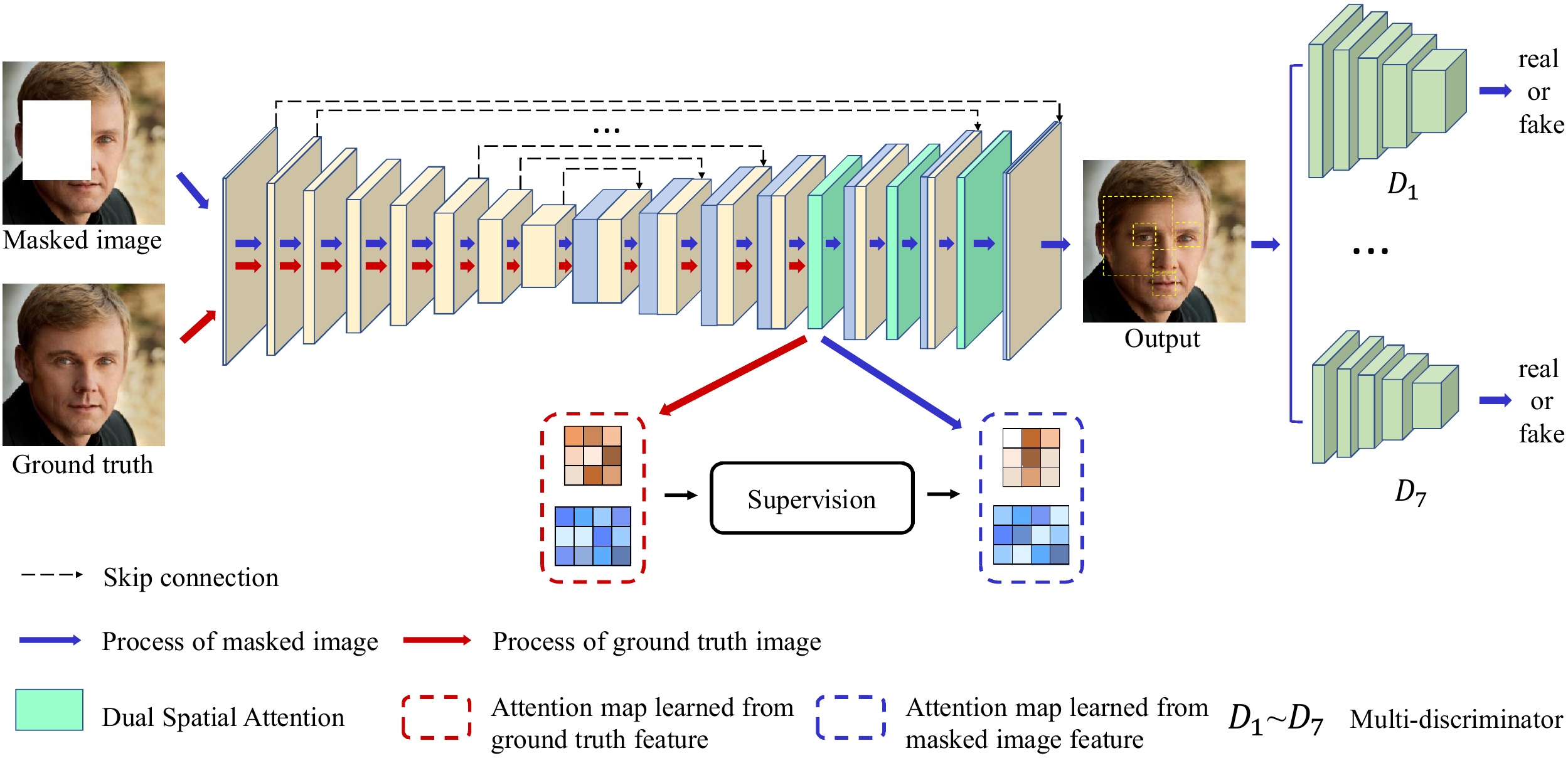}
\end{center}
   \caption{The overall architecture of our model. We use the U-Net structure~\cite{liu2018image} as backbone. The proposed attention module is embedded to layer 12, 13 and 14, where the resolution is $16 \times 16$, $32 \times 32$ and $64 \times 64$ respectively. We feed the ground truth images to the network, such that we can impose an oracle supervision signal on attention scores produced by DSA. We deploy discriminators on both the masked area and each of the four facial components, respectively. Best viewed in color.}
\label{fig:wholemodel}
\end{figure*}
\subsection{Image Inpainting}
Previous image inpainting methods can be divided into two categories: hand-crafted methods and learning-based methods.

Methods in the first category attempt to copy similar patches from the unmasked area to fill in the missing area~\cite{hongying2010image,ruvzic2014context,lee2012robust,he2012statistics}.
Criminisi~\etal~\cite{criminisi2004region} established the priority of each missing patch with reference to the surrounding structure. When the missing patch is surrounded by more valid pixels or closer to boundary region, this patch is assigned a higher priority. These methods can restore a continuous structure as they give priority to well-defined structural areas; however, iterating over every patch to search for the most similar one results in elevated time and memory costs. Accordingly, Barnes~\etal~\cite{barnes2009patchmatch} proposed a faster method referred to as PatchMatch. This approach employs a randomized algorithm to quickly find the approximate nearest neighbor matches. However, the hand-crafted methods are unable to handle complex-structured or largely occluded images, as only low-level features are considered.

The second category of methods~\cite{yeh2017semantic,ren2019structureflow,xie2019image} typically involve training a deep CNN in an encoder-decoder structure to predict each pixel of the missing area.
Pathak~\etal~\cite{pathak2016context} proposed Context Encoders that applied adversarial learning to the entire image.
In order to generate more realistic details, moreover, Iizuka~\etal~\cite{iizuka2017globally} appended an extra local discriminator to improve the generated effect of the masked area.
However, this approach relies on post-processing to alleviate artifacts. To address this problem, Liu~\etal~\cite{liu2018image} used only valid pixels per convolution to alleviate artifacts by updating a binary mask to indicate the generated pixels.

\textbf{Face Completion} Due to the complexity and diversity of facial structures, face completion is one of the more challenging image inpainting tasks. Generally speaking, researchers in this area therefore use a wealth of face prior knowledge to aid in restoration.
For example, Li~\etal~\cite{li2017generative} used face parsing to propose a semantic parsing loss.
Song~\etal~\cite{song2018geometry} trained an extra network to restore the facial landmarks and face parsing, then input them along with the masked image to train face completion network.
However, the results of these methods are greatly affected by the performance of the prior knowledge extracting network, which may consume a large amount of computational resources;
moreover, these methods cannot directly guide the network to focus on the texture of key facial components.
\subsection{Attention Mechanism}
In order to maintain contextual consistency, Yu~\etal~\cite{yu2018generative} proposed a coarse-to-fine network containing a contextual attention module that learns the correlations between the missing and unmasked patches.
Subsequently, some methods directly use the coarse-to-fine network with contextual attention; for example, GConv~\cite{yu2018free} turns the binary mask proposed by PConv~\cite{liu2018image} into learnable soft value as a gating mechanism, while Xiong~\etal~\cite{xiong2019foreground} used object contours as prior knowledge to assist in restoration.
On the other hand, other methods opt to use contextual attention in different ways. For example, Sagong~\etal~\cite{sagong2019pepsi} designed a parallel decoding network to replace the coarse-to-fine structure, thereby reducing the number of parameters, while Zeng~\etal~\cite{zeng2019learning} proposed an attention transfer network that uses the attention learned from high-level features to reconstruct low-level features. Moreover, the CSA layer~\cite{liu2019coherent} learns correlations between the patches inside the masked area.

In summary, previous attention-based methods learn long-range correlations in order to search the similar feature-level patches as references for filling. However, the learned attention is not reliable enough, as the parameters of the attention module lack direct supervision.
\section{Proposed Algorithm}
The overall architecture of the proposed approach is illustrated in Figure~\ref{fig:wholemodel}. We adopt the same U-Net structure as used in~\cite{liu2018image} to construct the basic generator. In what follows, we provide the details of our proposed methods. Specifically, we introduce the proposed DSA module with an oracle supervision signal, and then describe the deployment of the multi-discriminator. Finally, we describe the loss functions used to guide the training process.
\subsection{Dual Spatial Attention with Supervision}
\label{chap:DSA}
We treat the masked area as the foreground and the unmasked area as background. When learning relations between the different parts of the face, we consider two key scenarios. First, when restoring the foreground features, we obtain reference information from the background. For example, when the left eye is masked and the right one is not, we obtain features from the right eye to help restore the left eye. Second, when the masked area is large, we consider the relations between different parts in the foreground. For instance, when both eyes are masked, we ensure that the restored eyes have similar features.

Inspired by the principle of self-attention~\cite{zhang2019self,fu2019dual}, we propose the DSA module, which comprises foreground self-attention and foreground background cross-attention modules, so as to tackle above two scenarios.
\begin{figure}
\begin{center}
\includegraphics[width=1.0\linewidth]{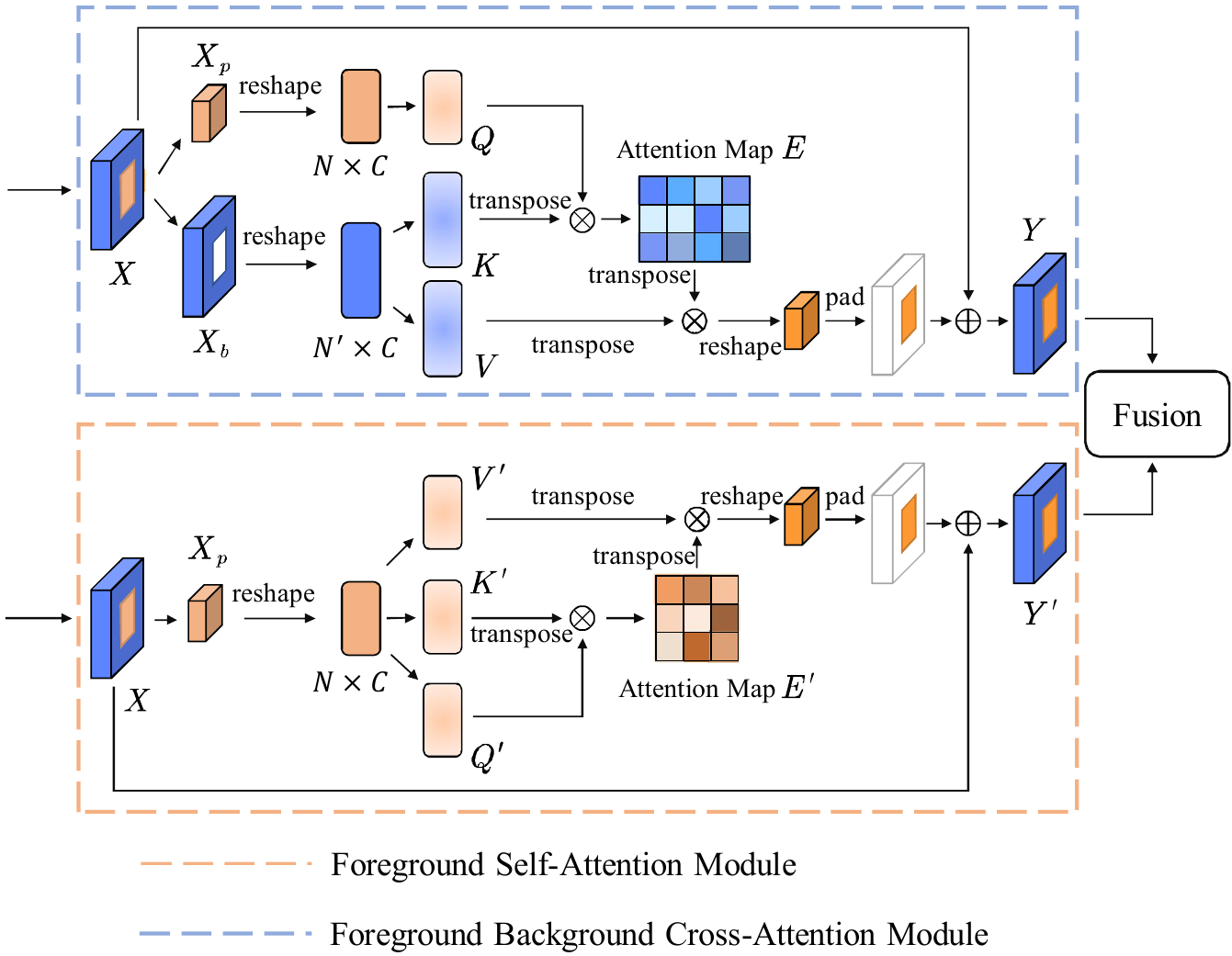}
\end{center}
   \caption{Overview of the Dual Spatial Attention (DSA) module, which includes two parallel branches. They focus on learning foreground-background cross-attention and foreground self-attention, respectively. Best viewed in color.}
\label{fig:attention}
\end{figure}
As shown in Figure~\ref{fig:attention}, we first use the mask to segment the input feature $\mathbf{X}$ into the foreground feature $\mathbf{X}_{p}$ and background feature $\mathbf{X}_{b}$. Then we reshape $\mathbf{X}_{p}$ into $N \times C$ and $\mathbf{X}_{b}$ into $N^{'} \times C$, where $N$ and $N^{'}$ denote the number of foreground pixels and background pixels in $\mathbf{X}$ respectively. $C$ denotes the number of channels.
For the foreground background cross-attention module, we put $\mathbf{X}_{p}$ and $\mathbf{X}_{b}$ into 1-dimensional convolutional layers respectively and generate two new feature maps $\mathbf{Q}$ and $\mathbf{K}$. Subsequently, we conduct a matrix multiplication between $\mathbf{Q}$ and the transpose of $\mathbf{K}$. Then we apply a softmax layer to obtain the attention map with size of $N \times N^{'}$. We write,
\begin{equation}
{E}_{i j}=\frac{\exp \left(\mathbf{Q}_{i} \cdot \mathbf{K}_{j}\right)}{\sum_{n=1}^{N^{'}} \exp \left( \mathbf{Q}_{i} \cdot \mathbf{K}_{n} \right)},
\end{equation}
where ${E}_{i j}$ denotes the degree of correlation between the $i^{th}$ feature vector of $\mathbf {Q}$ (learned from $\mathbf {X}_{p}$) and the $j^{th}$ feature vector of $\mathbf {K}$ (learned from $\mathbf {X}_{b}$).

Meanwhile, we also feed feature $\mathbf {X}_{b}$ into the other 1-dimensional convolutional layer, so as to generate a new feature $\mathbf {V}$. Next, we perform a matrix multiplication between the transpose of $\mathbf {V}$ and the transpose of the attention matrix, then reshape it back to the original size of $\mathbf {X}_{p}$ $\left(C \times H \times W \right)$. In this way, the original foreground features are eventually rebuilt with consideration given to the correlation with the background features. Finally, we extend the rebuilt feature to the original feature map size by means of zero-padding, and then merge it into the original feature map $X$ via element-wise sum operation, which can be formulated as follows:
\begin{equation}
\mathbf {Y} =\alpha \operatorname{Pad}\left(\mathbf {V}^{{\mathrm{T}}} \otimes \mathbf {E}^{{\mathrm{T}}}\right) + \mathbf {X},
\end{equation}
where $\operatorname{Pad}$ denotes zero-padding operation, $\alpha$ is a training parameter that is initialized to zero, and $\otimes$ denotes matrix multiplication.

For the foreground self-attention module, we follow the same steps as for the foreground background cross-attention module, except that we use $\mathbf {X}_{p}$ only. Specifically, we use three convolutional layers to generate feature maps $\mathbf {V}^{'}$, $\mathbf {K}^{'}$, and $\mathbf {Q}^{'}$. The foreground attention matrix can be formulated as
\begin{equation}
E^{'}_{i j}=\frac{\exp \left(\mathbf{Q}^{'}_{i} \cdot \mathbf{K}^{'}_{j}\right)}{\sum_{n=1}^{N} \exp \left(\mathbf {Q}^{'}_{i} \cdot
\mathbf{K}^{'}_{n}\right)},
\end{equation}
where $E^{'}_{i j}$ denotes the degree of correlation between the $i^{th}$ feature vector of $\mathbf {Q}^{'}$ and $j^{th}$ feature vector of $\mathbf {K}^{'}$, where $\mathbf {Q}^{'}$ and $\mathbf {K}^{'}$ are both learned from $\mathbf {X}_{p}$. The output can thus be formulated as follows,
\begin{equation}
\mathbf {Y}^{'} =\beta \operatorname{Pad}\left(\mathbf {V}^{'^{\mathrm{T}}} \otimes \mathbf {E}^{'^{\mathrm{T}}}\right) + \mathbf {X},
\end{equation}
where $\beta$ is a training parameter initialized to zero as $\alpha$.
Finally, we fuse $\mathbf {Y}$ and $\mathbf {Y}^{'}$ by means of element-wise addition, and then adopt a convolutional layer to adjust the feature and obtain the final refined feature map.

\textbf{Supervision Signal} \label{chap:kl}
The attention module helps the network select reference features to improve the filling quality. Nevertheless, if the learned attention is insufficiently accurate, the network may refer to unsuitable features, resulting in poorer filling quality. As a result, ensuring the accuracy of the learned attention scores is key to improving the filling quality. However, the parameters of the attention module are optimized together with the total face completion network using supervision signals for face completion only, meaning that the parameters of the attention module lack a direct restriction.

In order to impose a direct supervision signal towards attention, we extract the attention learned from the ground truth images as the objective. More specifically, in addition to taking the masked image as input, we also feed the ground truth image into the network when training. Through the use of the same network layers, including the DSA, we can obtain the oracle attention maps learned from the ground truth denoted as $\mathbf {E}^{gt}$ and $\mathbf {E}^{'^{gt}}$. Moreover, all attention scores are received through the softmax layer, which means that the attention map comprises a number of probability distributions. Therefore, we use KL-divergence distance to set up the objective function for attention. The KL-divergence loss is formulated as follows:
\begin{equation}
\begin{aligned}
\mathcal{L}_{\text {KL}}=&\frac{1}{N \cdot N^{'}} \sum_{i=1}^{N} \sum_{j=1}^{N^{'}} E_{i j}^{gt} \cdot \left(\ln E_{i j}^{gt} - \ln E_{i j} \right)\\
&+ \frac{1}{N \cdot N} \sum_{i=1}^{N} \sum_{j=1}^{N} E_{i j}^{'^{gt}} \cdot \left(\ln E_{i j}^{'^{gt}} - \ln E_{i j}^{'} \right),
\end{aligned}
\label{equ:KL}
\end{equation}
where we add the average KL-divergence distance between $\mathbf {E}^{gt}$ and $\mathbf {E}$, as well as between $\mathbf {E}^{'^{gt}}$ and $\mathbf {E}^{'}$.

In addition, attentions learned from different scale layers complement each other. On a high-resolution feature map, the attention reflects the relations between small-scale features, such as hair. Conversely, the attention on a low-resolution feature map reflects the relations between large-scale features, such as large-range aspects of facial structure. After accounting for the trade-off between computational efficiency and attention learning, we embed the DSA module into three layers of the decoder (\ie layers 12, 13, and 14) to assist the filling process.
It is important to note that the large-scale attention represents more large-range structural information and can affect later features; thus, we only impose the KL-divergence loss on layer 12 in order to strengthen the authenticity of the face structure, which can also benefit the subsequent attention learning.

\textbf{Discussion} Compared with the contextual attention layer~\cite{yu2018generative}, our DSA uses matrix multiplication rather than dividing the background patches as kernels for convolution, which significantly promotes efficiency.
In addition, the CSA layer~\cite{liu2019coherent} also learns the relations between foreground patches. However, it computes the similarity between two foreground patches in a iterative way; this slows down the filling speed, especially when the masked area is large. Moreover, neither contextual attention or CSA ensures the accuracy of learned attention scores, which leads to unsatisfying results.
\subsection{Multi-discriminator Design}
\begin{figure}
\begin{center}
\includegraphics[width=0.9\linewidth]{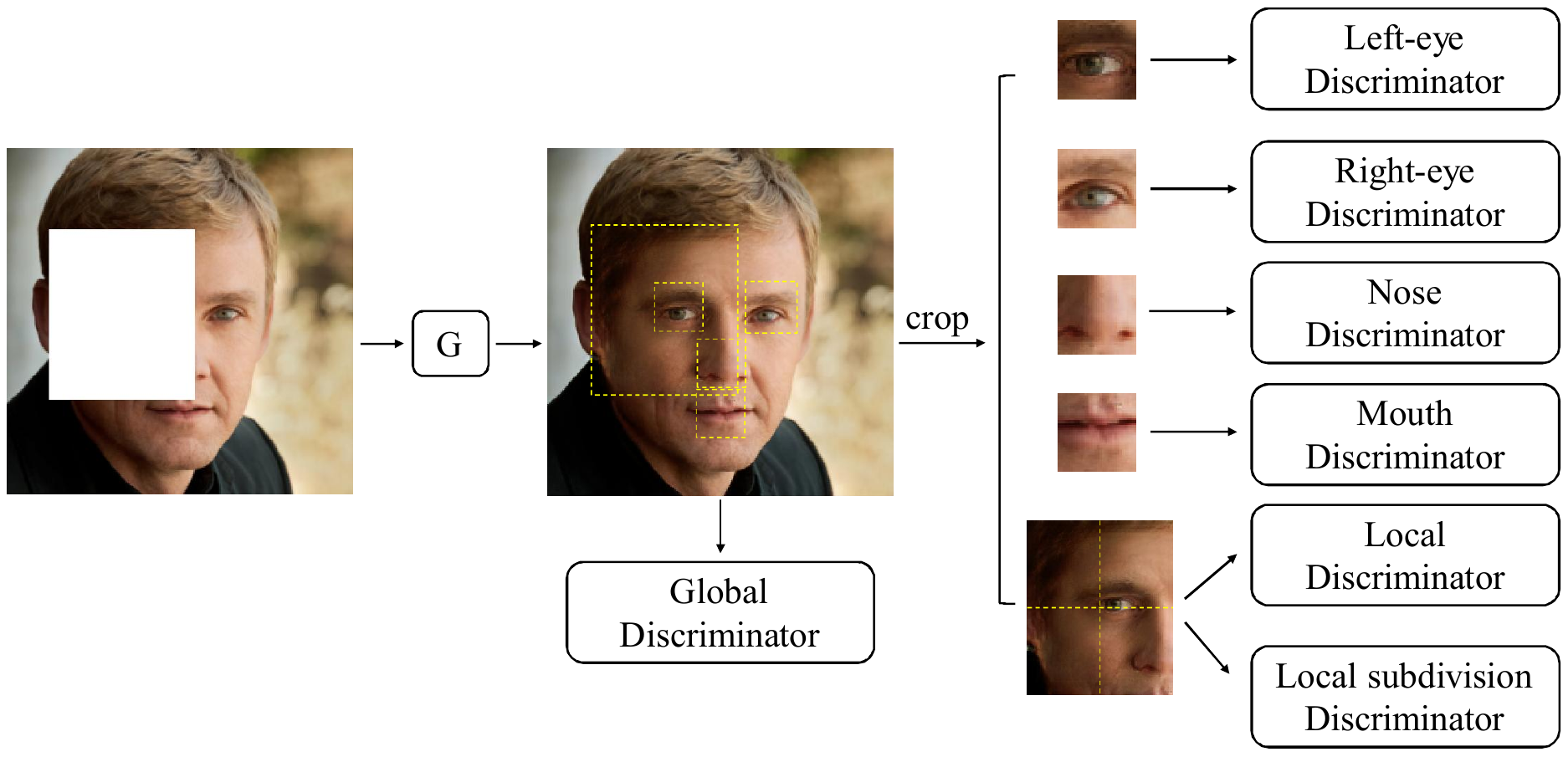}
\end{center}
   \caption{Overview of the multi-discriminator. $G$ denotes the generator. The four facial components are located by facial landmarks of the ground truth image. Best viewed in color.}
\label{fig:multiD}
\end{figure}

Adversarial learning is helpful in generating photo-realistic images by training the generator and discriminator until a Nash equilibrium is achieved. In addition to the global discriminator~\cite{pathak2016context}, which is used on the whole image, the local discriminator is designed to focus on the generated details of the masked area~\cite{iizuka2017globally, li2017generative}. For face completion tasks, the quality of the facial components largely determines the authenticity of the entire face image. However, relying solely on the global and local discriminators is insufficient if our goal is to guide the network to focus on small regions. To this end, we propose the multi-discriminator for enhancing face details especially on facial components, as shown in Figure~\ref{fig:multiD}.

We first use the facial landmarks extracted by the method~\cite{kazemi2014one} to mark the locations of the left eye, right eye, nose, and mouth. We then generate four masks of fixed size, and use these masks to crop four facial components. During training, we input each facial component of the generated image and ground truth into the corresponding discriminator to judge whether it is real or fake. Moreover, inspired by the collocation of the global and local discriminators, we further divide the masked area into four equal parts without overlap. These four parts share a discriminator, named the local subdivision discriminator, which focuses on more detailed characteristics and can also be regarded as a weighted optimization of the local discriminator loss function.

By using the multi-discriminator, the generator can learn more specific features for each facial component and further enhance the details within the masked area. Unlike previous ways that utilize the prior knowledge, we only use facial landmarks to mark the positions of facial components, after which we guide the network to improve the details in specific areas. Furthermore, since the discriminator only works during the training process, adding multiple discriminators does not affect the efficiency of the implementation.
\subsection{Loss Functions}
In order to effectively guide the training process, it is critical to design loss functions to be capable of measuring the distance between the generated image and the corresponding ground truth. We thus adopt multiple loss functions from different aspects, described as follows.

Given an original ground truth image $\mathbf {I}_{gt}$ and a randomly generated binary mask $\mathbf {M}_{1}$ (zero for holes), we produce the training image $\mathbf {I}_{in}$ by means of element-wise multiplication and denote the generated area of output $\mathbf {I}_{out}$ as $\mathbf {I}_{gen}$. Moreover, in order to crop the facial components, we also produce a left-eye mask, right-eye mask, nose mask, and mouth mask, which are denoted by $\mathbf {M}_{2},\ldots,\mathbf {M}_{5}$ respectively.

Firstly, we use L1-distance between the output and the ground truth as the reconstruction loss $\mathcal{L}_{r}$ to constrain the value of pixels. Furthermore, we propose to increase the penalty on the facial components and masked area. $\mathcal{L}_{r}$ can be represented as
\begin{equation}
\mathcal{L}_{r}=
\left\| ( \mathbf{1} + \sum_{i=1}^{5}\left(\mathbf{1}-\mathbf {M}_{i}\right))
\odot
\left( \mathbf {I}_{out}-\mathbf {I}_{gt} \right)\right\|_{1},
\end{equation}
where $\odot$ denotes element-wise multiplication, and $\mathbf{1}$ indicates a matrix with all values set to 1. Here, the facial components and missing area have more weight than other parts, while the missing portion of facial components is assigned the largest weight.
\begin{table}
\begin{center}
\begin{tabular}{lc c c c}
\hline
Method & L1 & PSNR & SSIM & LPIPS~\cite{zhang2018unreasonable} \\
\hline
PM~\cite{barnes2009patchmatch} & 5.82\% & 17.60 & 0.7786 & 0.2221 \\
CA~\cite{yu2018generative} & 1.82\% & 24.58 & 0.8980 & 0.0977 \\
PIC~\cite{zheng2019pluralistic} & 1.81\% & 25.31 & 0.9023 & 0.0897 \\
GConv~\cite{yu2018free} & 1.89\% & 26.29 & 0.8996 & 0.0809 \\
 \hline
Ours & \textbf{1.46\%} & \textbf{26.36} & \textbf{0.9107} & \textbf{0.0706} \\
\hline
\end{tabular}
\end{center}
\caption{Quantitative results on the same test set using masks with random position. Higher SSIM and PSNR values are better; lower L1 error and LPIPS are better.}
\label{table:comparison}
\end{table}
Secondly, we introduce the perceptual loss $L_{p}$ using a VGG-16~\cite{Simonyan2014Very}, which is pre-trained by~\cite{zhang2018unreasonable} to impose a constraint:
\begin{equation}
\begin{aligned}
\mathcal{L}_{p}=\left\|\Psi\left(\mathbf{I}_{o u t}\right)-\Psi\left(\mathbf{I}_{g t}\right)\right\|_{1}+
\left\|\Psi\left(\mathbf{I}_{gen}\right)-\Psi\left(\mathbf{I}_{g t}\right)\right\|_{1},
\end{aligned}
\end{equation}
where $\Psi$ is the output of the pre-trained VGG-16. The perceptual loss computes the L1-distance in feature space between both ${I}_{o u t}$ and ${I}_{gen}$ and the ground truth.

Thirdly, we adopt PatchGAN~\cite{isola2017image} as our discriminator structure, which maps the input image to a matrix where each element represents the authenticity of a portion of the input image. In this way, the network pays more attention to the local image details. In addition, we adopt an improved version of WGAN with a gradient penalty term~\cite{gulrajani2017improved}. The final adversarial loss function for each discriminator is as follows:
\begin{equation}
\begin{aligned}
\mathcal{L}_{D_{i}}&={\mathbb{E}_{\boldsymbol{\mathbf {I}_{out}} \sim \mathbb{P}_{g}}}
\left[D_{i}\left(C_{i}(\mathbf {I}_{out})\right)\right]\left.
\right.\\
&\left.
-\ {\mathbb{E}_{\boldsymbol{\mathbf {I}_{gt}} \sim \mathbb{P}_{data}}}\left[D_{i}\left(C_{i}(\mathbf {I}_{gt} )\right)\right]
\right.\\
&\left.
+\ \gamma {\mathbb{E}_{\hat{\mathbf {I}} \sim \mathbb{P}_{ \hat{\mathbf {I}}}}}
(\|\nabla_{\hat{\mathbf {I}}} D_{i}(C_{i}(\hat{\mathbf {I}}))\|_{2}-1)^{2},\right.
\end{aligned}
\end{equation}
where $D_{i}(i=1,\ldots,7)$ indicates one of the seven discriminators as shown in Figure~\ref{fig:multiD}. Here, $C_{i}$ denotes the crop operation used to obtain the corresponding areas from the image, $\hat{\mathbf {I}}$ is interpolated from pairs of points sampled from the real data distribution $\mathbb{P}_{data}$ and the generated distribution $\mathbb{P}_{g}$; $\nabla$ means the operation to compute gradient, and $\gamma$ is set to 10. Thus, the adversarial loss for the generator is as follows,
\begin{equation}
\mathcal{L}_{G}=-\sum_{i=1}^{7}\mathbb{E}_{\mathbf {I}_{out} \sim \mathbb{P}_{g}}
\left[D_{i}\left(C_{i}(\mathbf {I}_{out})\right)\right].
\end{equation}

The KL-divergence loss $\mathcal{L}_{KL}$ is depicted as Equation~\ref{equ:KL}, while the details are described in Section~\ref{chap:kl}. We define the overall loss function as follows,
\begin{equation}
\mathcal{L}= \lambda_{r}  \mathcal{L}_{r} + \lambda_{KL} \mathcal{L}_{KL} + \lambda_{p} \mathcal{L}_{p} + \lambda_{G} \mathcal{L}_{G},
\end{equation}
where we empirically set the four trade-off parameters $\lambda_{r}$, $\lambda_{KL}$, $\lambda_{p}$, and $\lambda_{G}$ to 10, 2, 1 and 1 respectively.
%-------------------------------------------------------------------------
\section{Experiment}
\begin{figure*}
\begin{center}
\includegraphics[width=.9\linewidth]{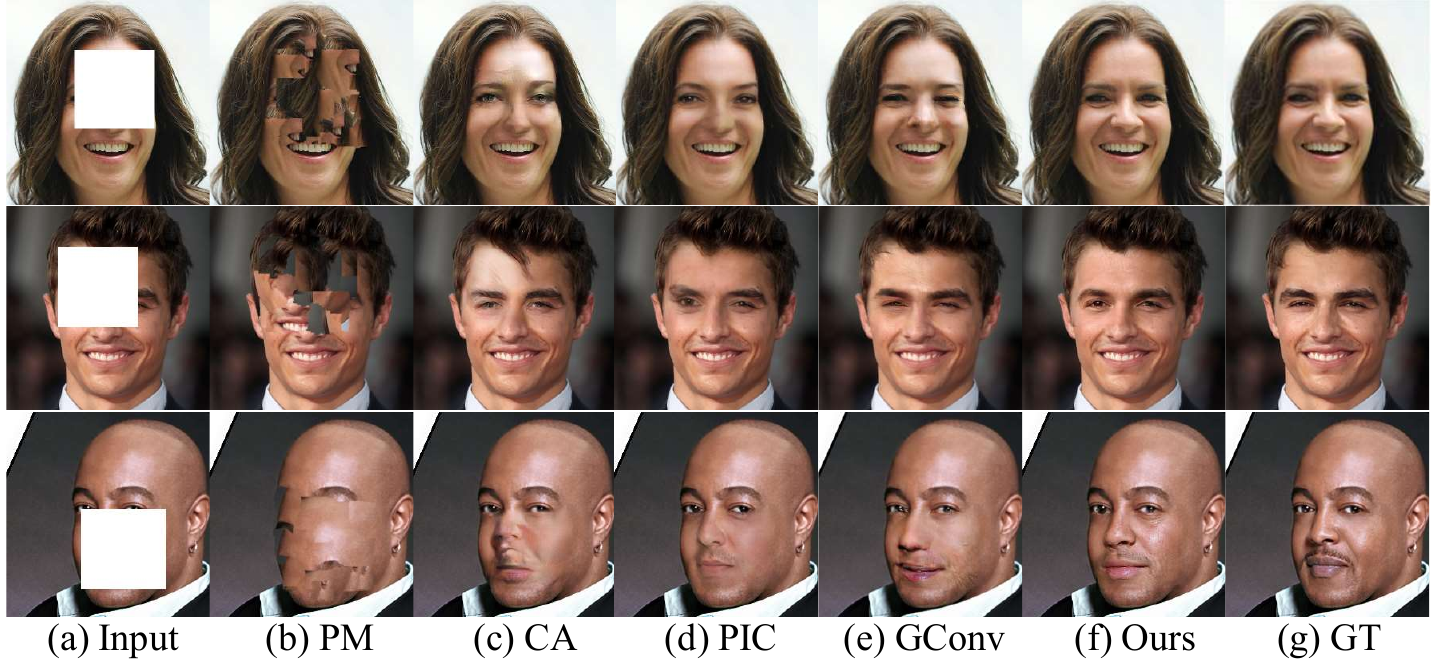}
\end{center}
   \caption{Comparisons in visual effect by different methods with random rectangle masks. Four state-of-the-art methods are compared: PM~\cite{barnes2009patchmatch}, CA~\cite{yu2018generative}, PIC~\cite{zheng2019pluralistic} and GConv~\cite{yu2018free}. Best viewed with zoom-in and pay attention to the details on facial components. More qualitative results are presented in the supplementary material.}
\label{fig:comparison1}
\end{figure*}
\subsection{Experimental settings}
\textbf{Dataset} We conduct a number of experiments on two high-quality human face datasets including CelebA-HQ~\cite{Karras2017Progressive} and Flickr-Faces-HQ~\cite{karras2019style}; these contain 30,000 and 70,000 high-quality face images respectively with size of $1024 \times 1024$. We randomly selected 2,000 images from CelebA-HQ and 10,000 from Flickr-Faces-HQ to comprise the test set.

\textbf{Implementation Details} We resize all input images to $256 \times 256$. The random position rectangle masks account for $13.5\% \sim 25\%$ of the original image; the largest size is $128 \times 128$ and the smallest size is $94 \times 94$. During training, we use RMSprop as the optimizer with a learning rate of 0.0001. On a single NVIDIA TITAN Xp (12GB), we train our model on CelebA-HQ for four days and Flickr-Faces-HQ for eight days with a batch size of 16.

\subsection{Comparisons with State-of-the-Art Methods}
\begin{table}
\begin{center}
\begin{tabular}{lc c c c}
\hline
Method & L1 & PSNR & SSIM & LPIPS~\cite{zhang2018unreasonable} \\
\hline
PEN-Net~\cite{zeng2019learning} & 2.87\% & 24.53 & 0.8369 & 0.1701  \\
\hline
Ours & \textbf{2.28\%} & \textbf{26.11} & \textbf{0.8718} & \textbf{0.1355} \\
\hline
\end{tabular}
\end{center}
\caption{Quantitative comparison on the same test set with a center mask. Higher SSIM and PSNR values are better; lower L1 error and LPIPS are better. }
\label{table:comparisonpen}
\end{table}
We conduct qualitative and quantitative comparisons with multiple methods including PatchMatch (PM)~\cite{barnes2009patchmatch}, Contexual Attention (CA)~\cite{yu2018generative}, PIC~\cite{zheng2019pluralistic}, GConv~\cite{yu2018free} and PEN-Net~\cite{zeng2019learning}. We use the officially released models of CA, PIC, and GConv trained on CelebA-HQ to facilitate fair comparisons. We compare with PEN-Net on the same test set on CelebA-HQ as~\cite{zeng2019learning}. As PEN-Net only handles a center mask on CelebA-HQ whose size is $128 \times 128$, we compare with it separately using a center mask. For PIC, we follow the official instructions to select the best of their multiple results. In addition, we train models of CA, PIC, and GConv for Flickr-Faces-HQ respectively. The results on Flickr-Faces-HQ are presented in the supplementary material. In order to focus on assessing the generative capability of the different models, we copy the valid pixels to the output image for all models in comparison.

\textbf{Quantitative Comparisons} As outlined in Table~\ref{table:comparison} and Table~\ref{table:comparisonpen}, we conduct quantitative comparisons on CelebA-HQ using different masks. We select the commonly-used L1 loss, peak signal-to-noise ratio (PSNR) and structural similarity (SSIM) as evaluation metrics at pixel space. As mentioned in~\cite{yu2018generative,molodetskikh2019perceptually}, however, these classical metrics are not optimal for image inpainting tasks; thus, we further use Learned Perceptual Image Patch Similarity (LPIPS)~\cite{zhang2018unreasonable} as perceptual metric. According to these metrics, our method not only outperforms the previous state-of-the-art image inpainting methods, but also achieves a notable improvement. This is because our method focuses on learning the exact relations of the facial structure and improving the completion effect of facial components; by contrast, other methods tend to ignore the complexity and diversity of the facial structure.
\begin{table}
\begin{center}
\begin{tabular}{l|c|c|c|c|c}
\hline
Method & L1 & PSNR & SSIM & LPIPS & Time Cost \\
 \hline
CA~\cite{yu2018generative} & 1.64\% & 25.56 & 0.9057 & 0.0899 & 12.9ms \\
DSA & \textbf{1.46\%} & \textbf{26.36} & \textbf{0.9107} & \textbf{0.0706} & \textbf{6.4ms} \\
\hline
\end{tabular}
\end{center}
\caption{Quantitative comparison results on CelebA-HQ. The first row is the result of our model that replaces DSA with the CA module~\cite{yu2018generative}. Higher SSIM and PSNR values are better; lower L1 error and LPIPS are better. We compare the time cost per image by CA and DSA on layer 14 of the U-Net.}
\label{table:ablation1}
\end{table}
\begin{figure}
\begin{center}
%\fbox{\rule{0pt}{2in} \rule{.9\linewidth}{0pt}}
\includegraphics[width=.95\linewidth]{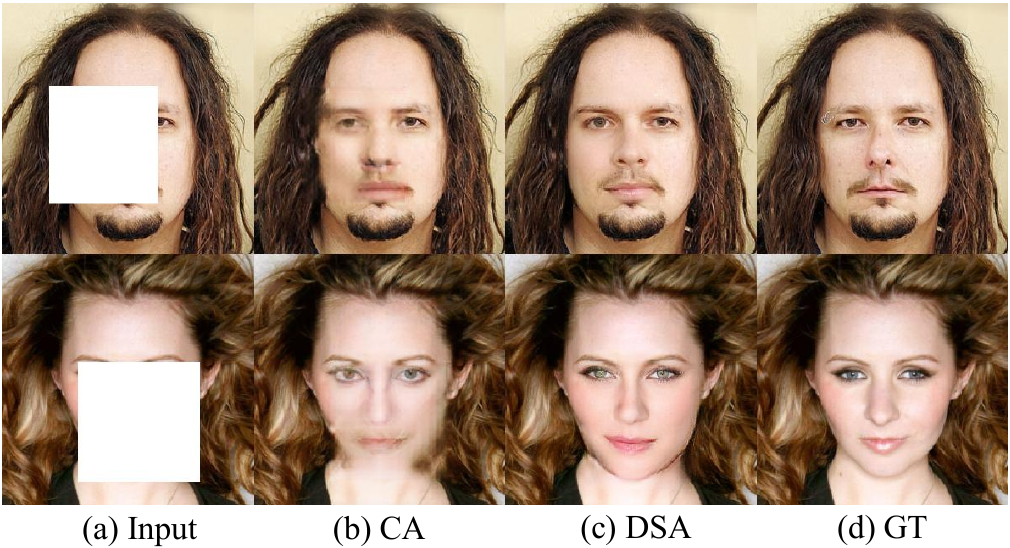}
\end{center}
   \caption{Qualitative comparison results on CelebA-HQ. In each row from left to right: (a) input image, (b) the result of using CA module~\cite{yu2018generative} to replace DSA layer, (c) the result of using DSA module and (d) the ground truth image. Best viewed with zoom-in.}
\label{fig:ablation1}
\end{figure}

\textbf{Qualitative Comparisons} As shown in Figure~\ref{fig:first} and Figure~\ref{fig:comparison1}, we respectively compare the visual effect using a center mask and a random mask on the CelebA-HQ dataset.
Firstly, the hand-crafted method PM~\cite{barnes2009patchmatch} fails to recover the basic structure of the face because it is difficult to find similar patches on background.
Secondly, CA~\cite{yu2018generative}, PIC~\cite{zheng2019pluralistic}, GConv~\cite{yu2018free} and PEN-Net~\cite{zeng2019learning} employ attention modules to learn contextual information for inpainting. But they still generate semantically inconsistent structures or textures as their learned attention scores may not be reasonable. In addition, these methods may generate artifacts or blurry effect on facial components such as the noses in the first row of Figure~\ref{fig:comparison1}. This is because they do not pay enough attention to facial components.
Finally, our approach produces sharper and more abundant details, especially for facial components, which can be explained by the effect of multi-discriminator.
More qualitative results are presented in the supplementary material.
\begin{figure}
\begin{center}
%\fbox{\rule{0pt}{2in} \rule{.9\linewidth}{0pt}}
\includegraphics[width=.95\linewidth]{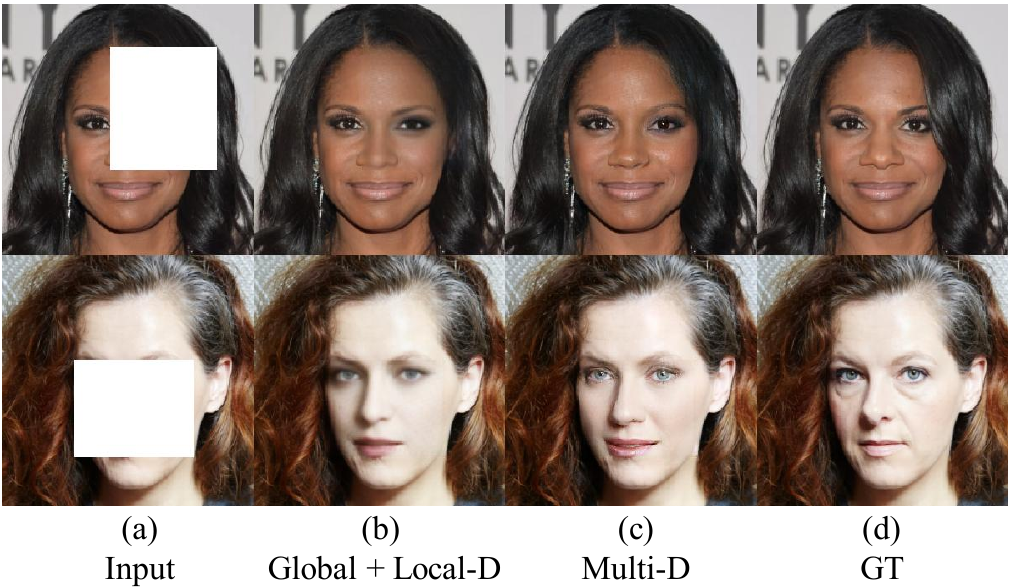}
\end{center}
   \caption{Qualitative comparison results on CelebA-HQ. In each row from left to right: (a) input image, (b) the result of using global and local discriminator, (c) the result of using the multi-discriminator, and (d) the ground truth image. Best viewed with zoom-in.}
\label{fig:ablation2}
\end{figure}
\subsection{Ablation Study}
\textbf{Effect of DSA Module} We replace the DSA module in our model with the contextual attention (CA) module~\cite{yu2018generative} for comparison.
As shown in Table~\ref{table:ablation1} and Figure~\ref{fig:ablation1}, the model with DSA is more effective and produces images with fine-grained textures. This is because DSA explores pixel-to-pixel correlations rather than patch-to-patch correlations in CA.
Therefore, DSA can employ small-scale features as reference for inpainting.
Furthermore, as DSA learns foreground self-attention, it maintains the contextual consistency in foreground areas. For example, the eyes generated by DSA are symmetric in both structure and texture, as shown in Figure~\ref{fig:ablation1}. In comparison, the eyes generated by CA may be inconsistent with each other. Besides, DSA is also more efficient than CA as the latter one adopts more convolution operations~\cite{yu2018generative}. The CSA module~\cite{liu2019coherent} also learns correlations between foreground patches, but its computational cost is significantly higher than DSA. This is because it adopts an iterative processing strategy, which means foreground pixels are refined one by one. Under the same GPU and experimentation setting, DSA and CSA (our implementation) take 6.4 ms and 31.3 ms per image on layer 14, respectively.

\textbf{Effect of Multi-discriminator} As shown in Figure~\ref{fig:ablation2}, comparisons in visual effect show that using global and local discriminator only~\cite{li2017generative, iizuka2017globally} is insufficient to generate high-fidelity facial textures. With the help of local subdivision discriminator and four discriminators for facial components, the generated results are sharper and contain richer textures especially for the facial components.
\begin{figure}
\begin{center}
%\fbox{\rule{0pt}{2in} \rule{.9\linewidth}{0pt}}
\includegraphics[width=.95\linewidth]{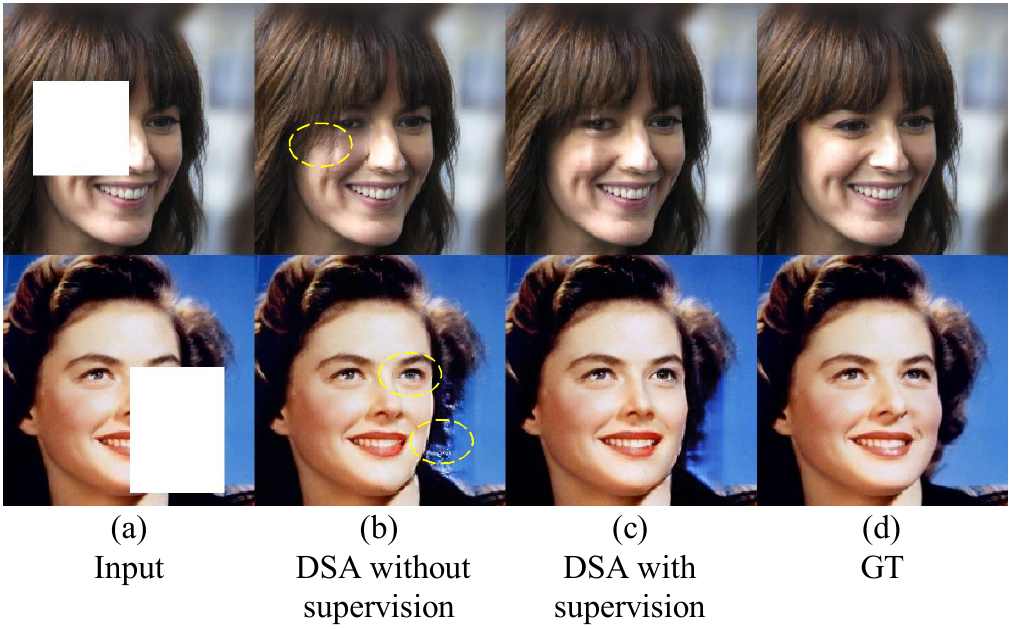}
\end{center}
   \caption{Qualitative comparison results on CelebA-HQ. In each row from left to right: (a) input, (b) the result of using DSA module but without supervision signal, (c) the result of using DSA module with supervision signal and (d) the ground truth image. The yellow circles indicate artifacts. Best viewed with zoom-in.}
\label{fig:ablation3}
\end{figure}
\begin{table}
\begin{center}
\begin{tabular}{l|c|c|c}
\hline
Method & PSNR & SSIM & LPIPS \\
 \hline
DSA without supervision & 26.23 & 0.9062 & 0.0726 \\
DSA with supervision    & \textbf{26.36} & \textbf{0.9107} & \textbf{0.0706} \\
\hline
\end{tabular}
\end{center}
\caption{Quantitative results over CelebA-HQ to prove the effect of supervision on DSA. Higher SSIM and PSNR values are better; lower LPIPS are better.}
\label{table:ablation3}
\end{table}

\textbf{Effect of Supervision on Attention} As shown in Figure~\ref{fig:ablation3} (b), DSA without the extra supervision signal may produce artifacts for some challenging cases. For example, redundant hair is generated on the cheek of the girl in the first row. In the second row, the generated eye is close to blue due to the influence of the blue background. In addition, there is clutter on the boundary between hair and background.
These artifacts are caused by inaccurate attention scores, which lead the network to refer to features in unsuitable areas for inpainting. As illustrated in Figure~\ref{fig:ablation3} (c), after imposing the oracle supervision signal as a guidance, the accuracy of attention scores is improved and accordingly these artifacts or color discrepancy can be solved. As noted in Table~\ref{table:ablation3}, the quantitative results also prove that the network benefits from the supervision signal.
%------------------------------------------------------------------------
\section{Discussion and Extension}
In this paper, we propose a comprehensive model for face completion consisting of a DSA module with supervision signal, and a multi-discriminator, and further conduct multiple experiments to prove that it outperforms previous state-of-the-art methods. The supervised DSA module helps the network to identify the correlation between different facial parts, while the multi-discriminator forces the network to learn the specific features of facial components.
In the supplementary material, we further show the results of the proposed method conditioning on irregular masks and higher resolution face image ($1024 \times 1024$).
%We subsequently use the proposed method to conduct high-resolution face completion ($1024 \times 1024$) and present the results in the supplementary material.
\section{Acknowledgement}
Changxing Ding is supported by NSF of China under Grant 61702193 and U1801262, the Science and Technology Program of Guangzhou under Grant 201804010272, the Program for Guangdong Introducing Innovative and Entrepreneurial Teams under Grant 2017ZT07X183, and the Fundamental Research Funds for the Central Universities of China under Grant 2019JQ01.

%\subsection{References}
{\small
\bibliographystyle{ieee_fullname}
\bibliography{egbib}
}

%\end{document}
\clearpage
\appendix
\section*{Appendix}
This supplementary material includes five sections.
Section~\ref{chp:sec1} shows comparison results between our method and state-of-the-art methods on the Flickr-Faces-HQ~\cite{karras2019style} database.
Section~\ref{sec2} provides more qualitative comparions on the CelebA-HQ~\cite{Karras2017Progressive} database.
Section\ref{sec6} shows two challenging cases including faces in profile or complex illuminations.
Section~\ref{sec3} shows face completion results of our method using irregular masks.
Section~\ref{sec4} provides the face completion results of our approach on high resolution images ($1024 \times 1024$). Finally, we introduce the details of the network architecture in Section~\ref{sec5}.

\section{Comparisons on Flickr-Faces-HQ}
\label{chp:sec1}
%FFHQ
\begin{table} [htb]
\begin{center}
\begin{tabular}{lc c c c}
\hline
Method & L1 & PSNR & SSIM & LPIPS~\cite{zhang2018unreasonable} \\
\hline
CA~\cite{yu2018generative} & 1.96\% & 24.30 & 0.8896 & 0.0869 \\
PIC~\cite{zheng2019pluralistic} & 1.88\% & 24.53 & 0.9007 & 0.0982 \\
GConv~\cite{yu2018free} & 1.85\% & 24.93 & 0.8879 & 0.0879 \\ %Image_NEAREST
 \hline
Ours & \textbf{1.50\%} & \textbf{26.06} & \textbf{0.9045} & \textbf{0.0693} \\
\hline
\end{tabular}
\end{center}
\caption{Quantitative comparisons on the same test set using rectangular masks of random position. Higher SSIM and PSNR values are better; lower L1 error and LPIPS values are better.}
\label{table:comparison_ffhq}
\end{table}

\begin{figure*}
\begin{center}
%\fbox{\rule{0pt}{2in} \rule{.7\linewidth}{0pt}}
\includegraphics[width=.9\linewidth]{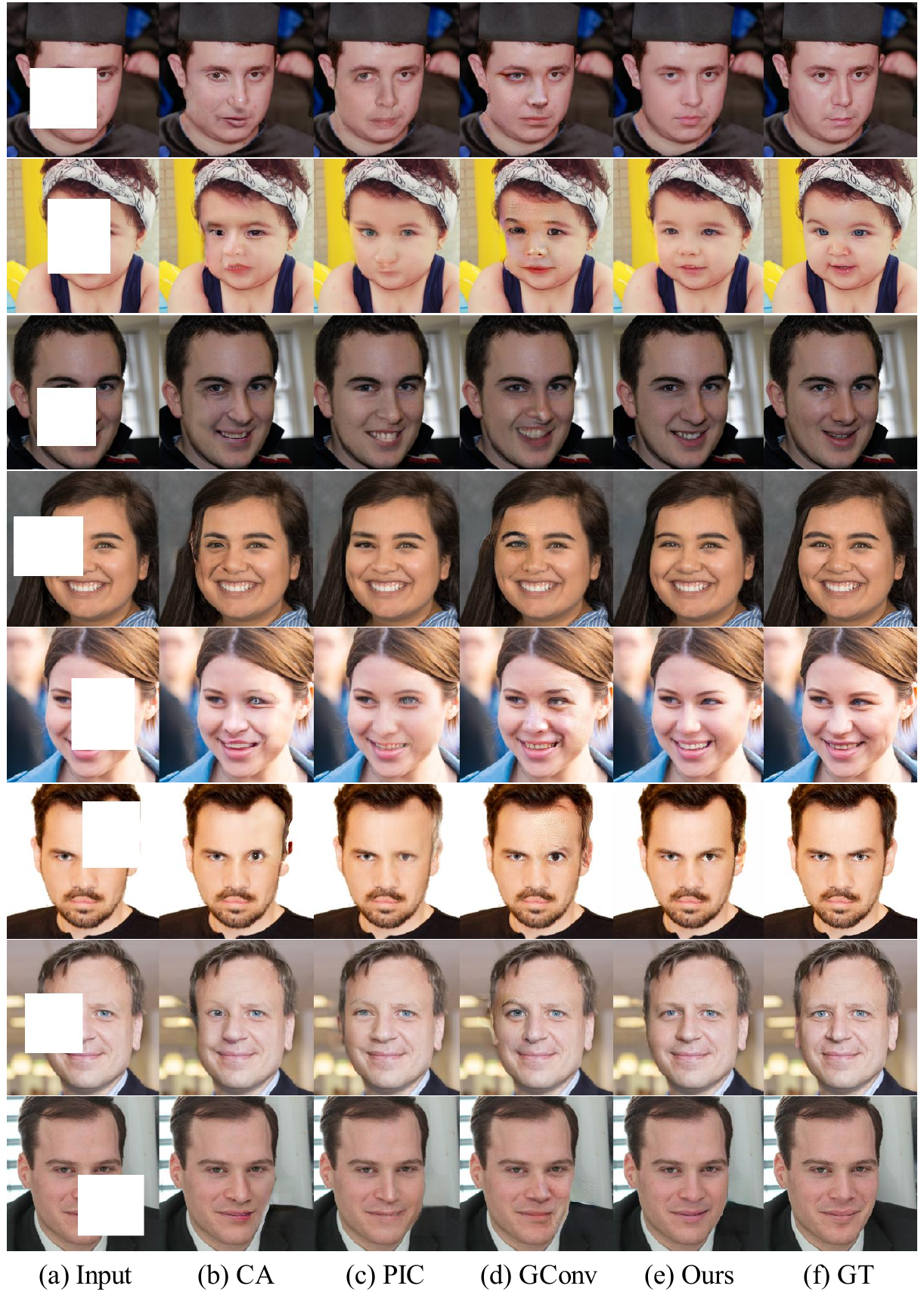}
\end{center}
   \caption{Comparisons on Flickr-Faces-HQ~\cite{karras2019style} by different methods with random rectangular masks. Three state-of-the-art methods are compared: CA~\cite{yu2018generative}, PIC~\cite{zheng2019pluralistic} and GConv~\cite{yu2018free}. Best viewed with zoom-in and pay attention to the details on facial components.}
\label{fig:ffhq}
\end{figure*}

We conduct both quantitative and qualitative comparisons between our approach and state-of-the-art methods on the Flickr-Faces-HQ~\cite{karras2019style} database. Rectangular masks of random position are adopted. All images are resized to $256 \times 256$. As the original papers of CA~\cite{yu2018generative}, PIC~\cite{zheng2019pluralistic}, and GConv~\cite{yu2018free} do not provide the performance of their models on Flickr-Faces-HQ, we use their released codes to train the three models on Flickr-Faces-HQ respectively.
Quantitative comparison results are summarized in Table~\ref{table:comparison_ffhq}, it is shown that our approach outperforms the other methods by large margins. Qualitative comparisons are provided in Figure~\ref{fig:ffhq}, it is clear that our method is also the best in visual effect.

\section{More Comparison Results on CelebA-HQ}
\label{sec2}
\begin{figure*}
\begin{center}
%\fbox{\rule{0pt}{2in} \rule{.7\linewidth}{0pt}}
\includegraphics[width=.9\linewidth]{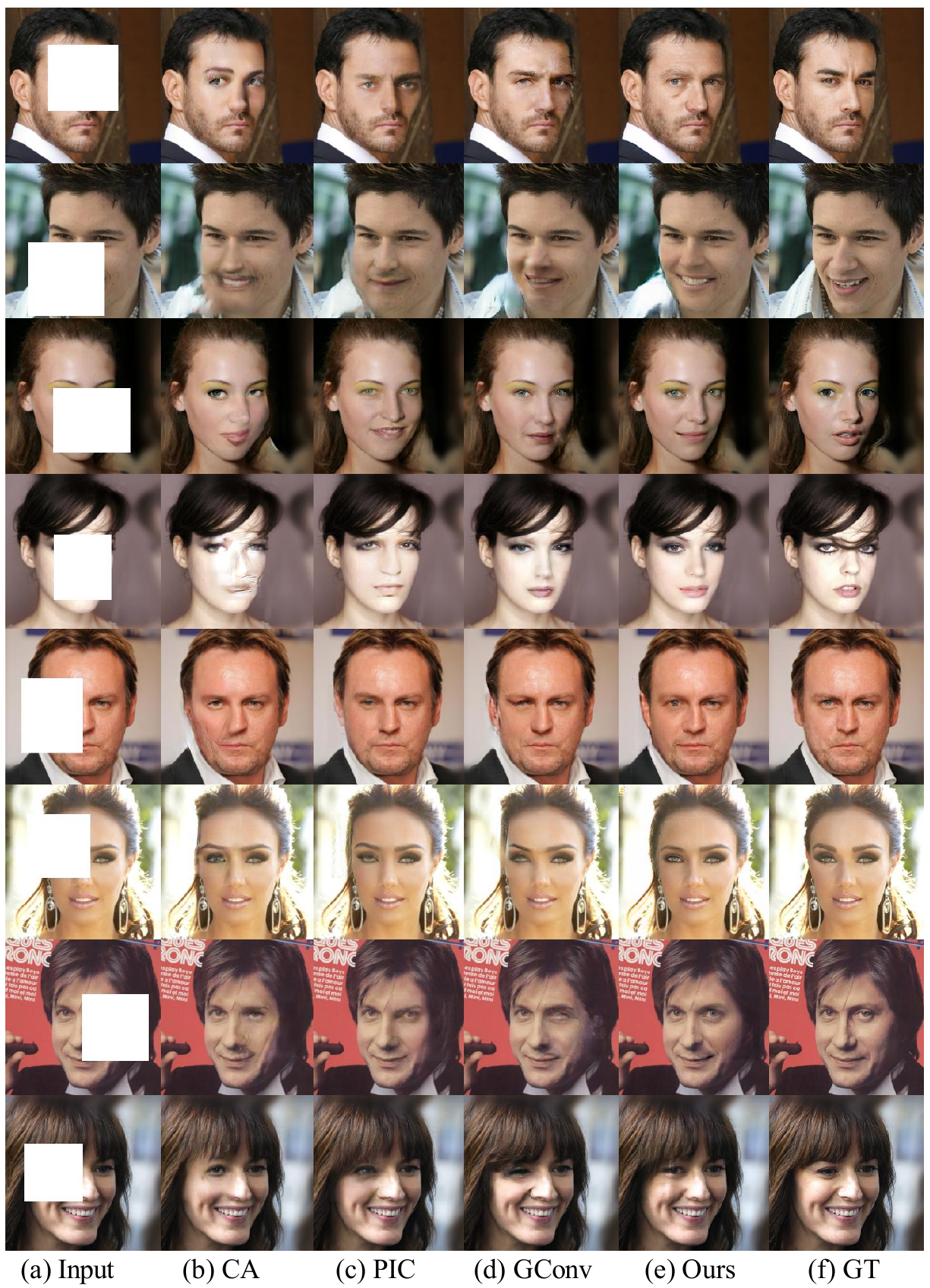}
\end{center}
   \caption{Comparisons on CelebA-HQ~\cite{Karras2017Progressive} by different methods with random rectangular mask. Three state-of-the-art methods are compared: CA~\cite{yu2018generative}, PIC~\cite{zheng2019pluralistic} and GConv~\cite{yu2018free}. Best viewed with zoom-in and pay attention to the details on facial components.}
\label{fig:celebRandom}
\end{figure*}

\begin{figure*}
\begin{center}
%\fbox{\rule{0pt}{2in} \rule{.7\linewidth}{0pt}}
\includegraphics[width=.9\linewidth]{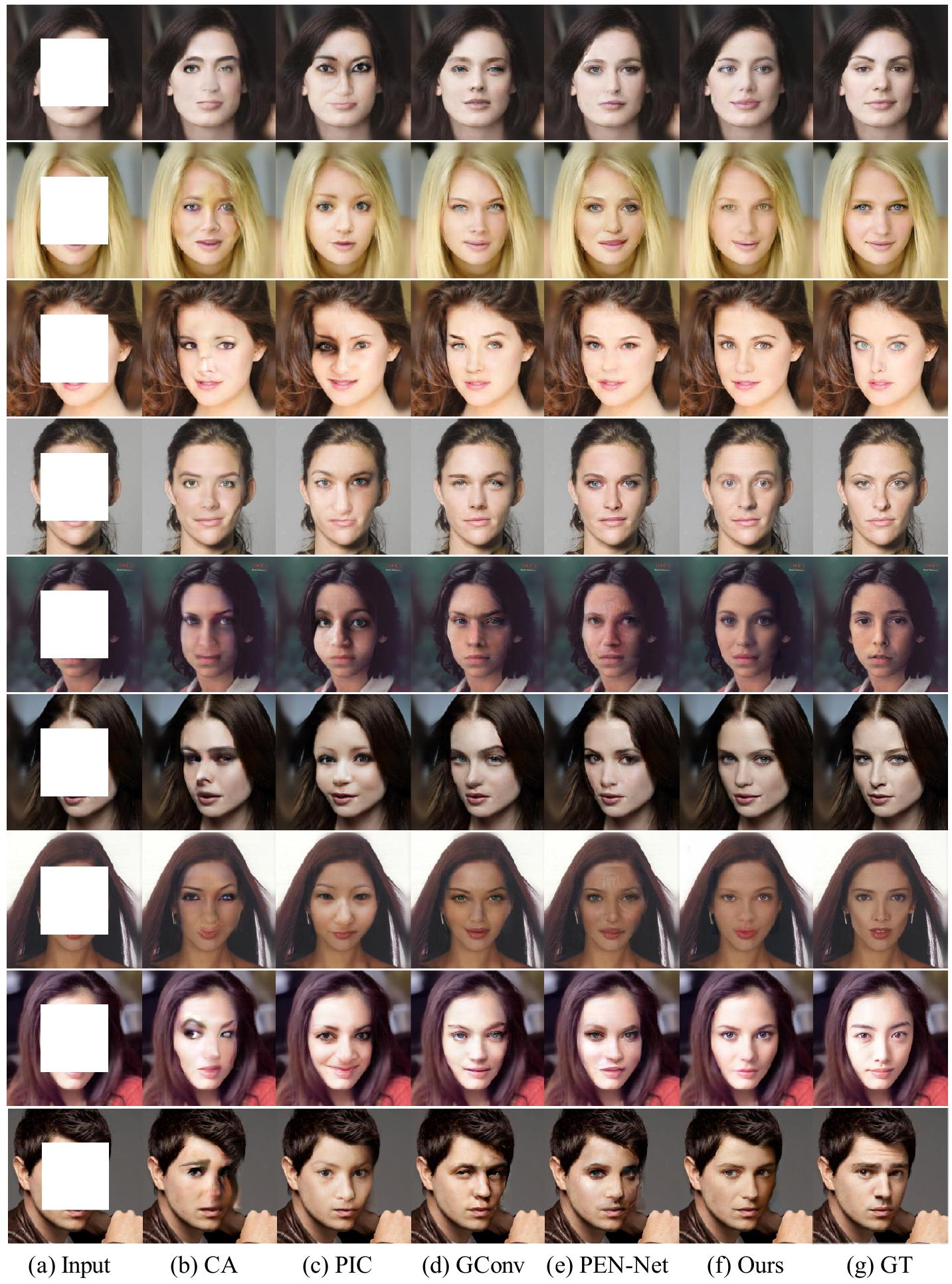}
\end{center}
   \caption{Comparisons on CelebA-HQ~\cite{Karras2017Progressive} by different methods with a center mask ($128 \times 128$). Four state-of-the-art methods are compared: CA~\cite{yu2018generative}, PIC~\cite{zheng2019pluralistic}, GConv~\cite{yu2018free} and PEN-Net~\cite{zeng2019learning}. Best viewed with zoom-in and pay attention to the details on facial components.}
\label{fig:celebCenter}
\end{figure*}

We show more qualitative comparisons on the CelebA-HQ~\cite{Karras2017Progressive} database in Figure~\ref{fig:celebRandom} and Figure~\ref{fig:celebCenter}.
In the two figures, rectangular masks of random position and a center mask are utilized, respectively. Intuitively, the center mask is more challenging as most facial components are occluded and it is difficult to find references from the background. It is shown that our method performs the best in visual effect.
\section{Special Cases}%
\begin{figure*}
\begin{center}
%\fbox{\rule{0pt}{2in} \rule{.7\linewidth}{0pt}}
\includegraphics[width=.9\linewidth]{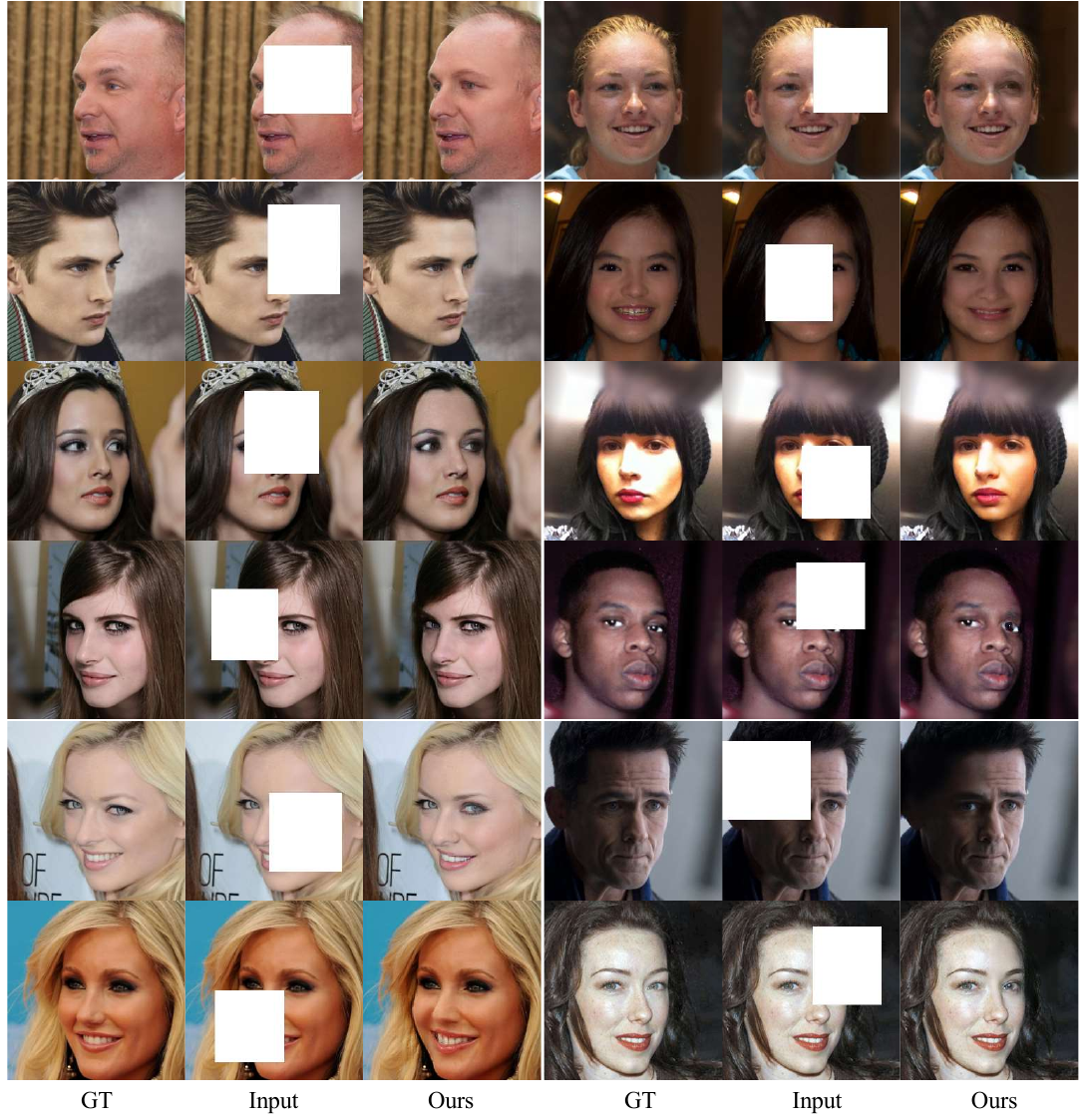}
\end{center}
   \caption{The results of processing faces in profile or complex illuminations. All these images are included in the CelebA-HQ test set. Best viewed with zoom-in.}
\label{fig:rebuttal}
\end{figure*}

\label{sec6}
As shown in Figure~\ref{fig:rebuttal}, we show the results of processing faces in profile or complex illuminations, which are indeed more challenging for the inpainting task due to the data imbalance problem.
\section{Results with Irregular Masks}
\label{sec3}
\begin{figure*}
\begin{center}
%\fbox{\rule{0pt}{2in} \rule{.7\linewidth}{0pt}}
\includegraphics[width=.9\linewidth]{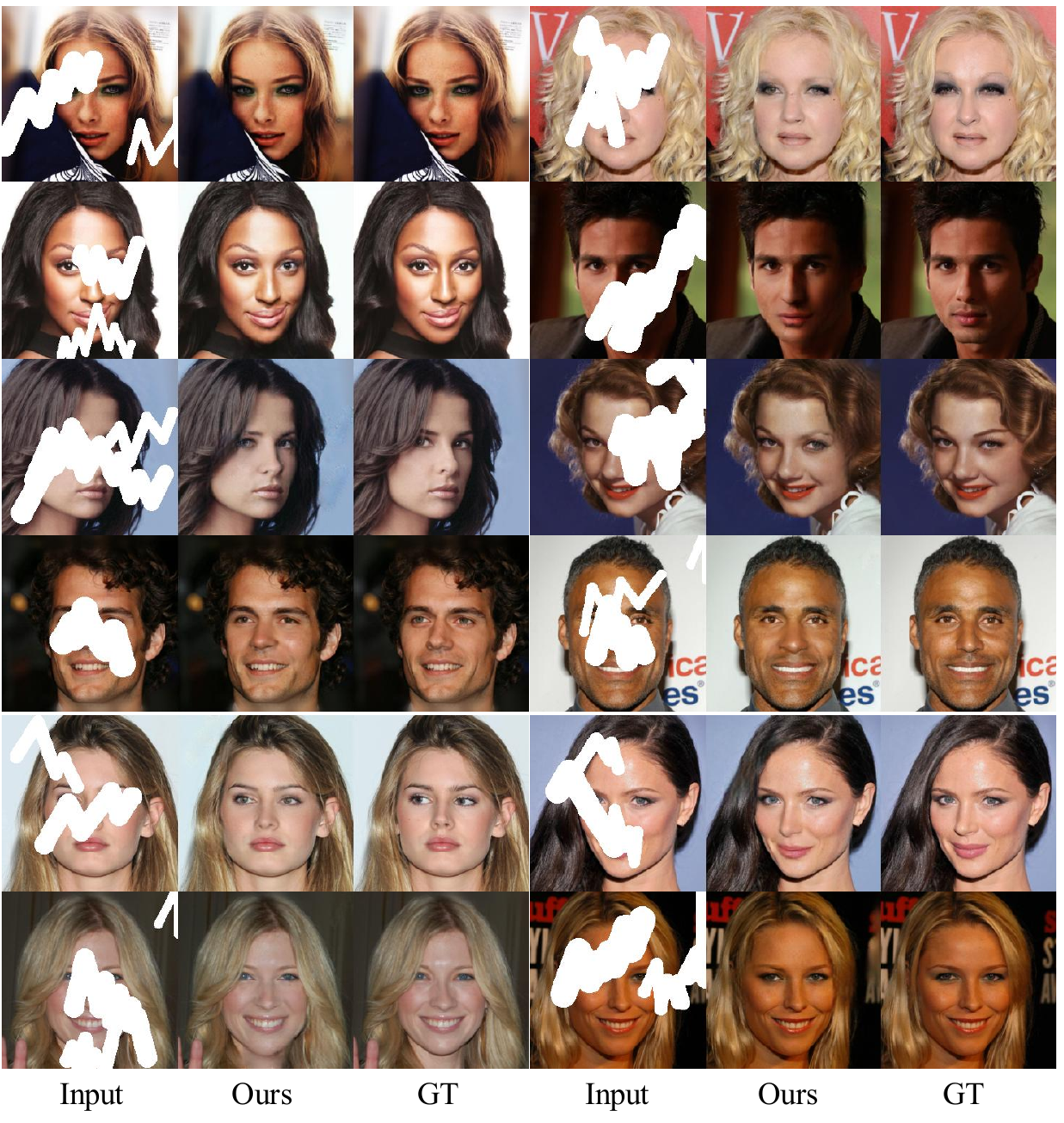}
\end{center}
   \caption{Results on CelebA-HQ with irregular masks.}
\label{fig:irregular}
\end{figure*}

All the above experiments adopt rectangular masks. In this experiment, we show face completion results of our approach using irregular masks. As the masks are irregular now and discriminators usually require rectangular patches as input, we consistently apply the local discriminator and local subdivision discriminator to the central region ($128 \times 128$) of each training image.
The other settings of our approach remain unchanged.
Face completion results are illustrated in Figure~\ref{fig:irregular}. It is shown that our approach produces face images of high-fidelity even with irregular masks.
\section{Results on High Resolution Images}
\label{sec4}
\begin{figure*}
\begin{center}
%\fbox{\rule{0pt}{2in} \rule{.7\linewidth}{0pt}}
\includegraphics[width=.9\linewidth]{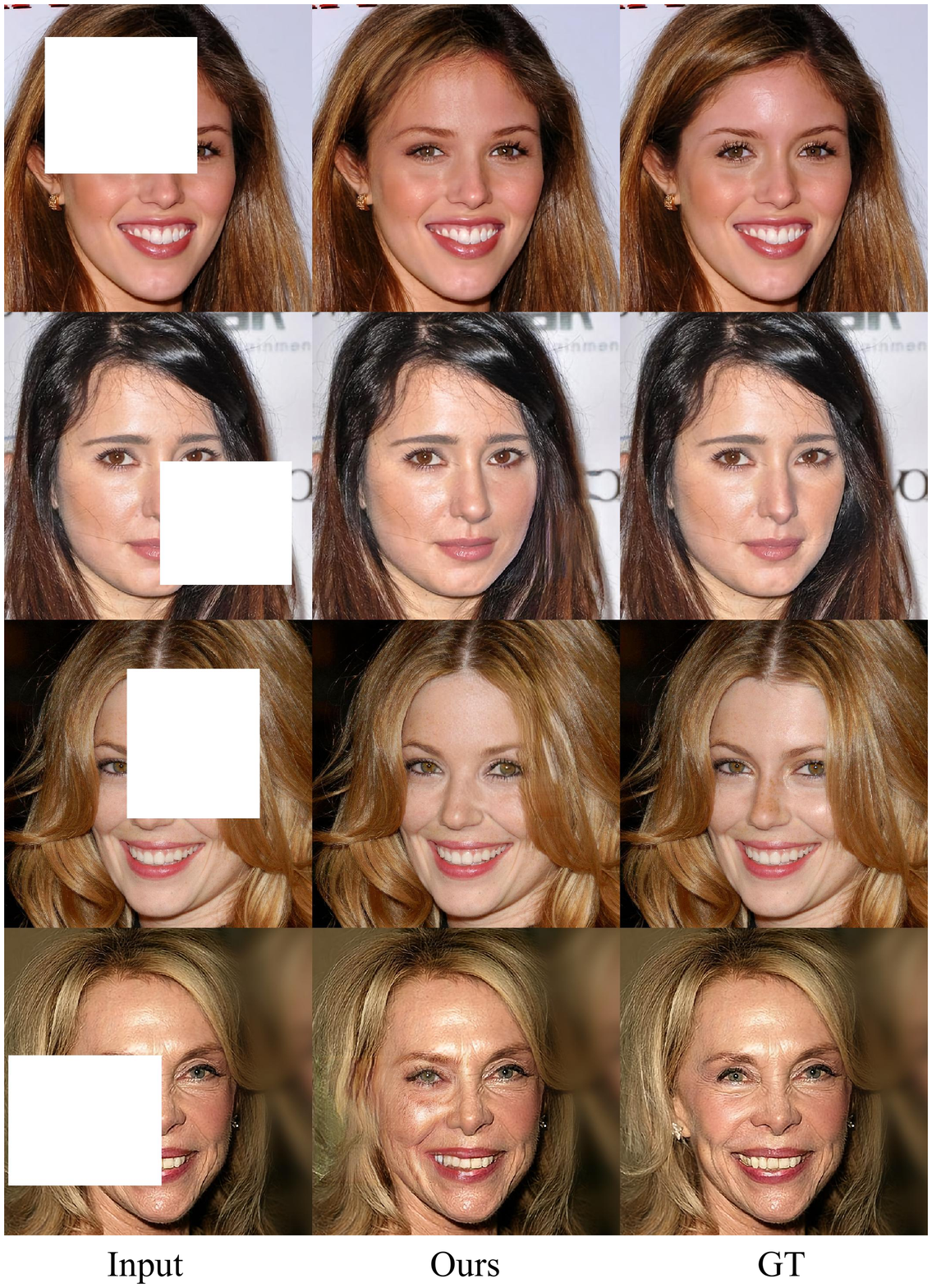}
\end{center}
   \caption{Results on high-resolution images of CelebA-HQ ($1024 \times 1024$).}
\label{fig:10241}
\end{figure*}

\begin{figure*}
\begin{center}
%\fbox{\rule{0pt}{2in} \rule{.7\linewidth}{0pt}}
\includegraphics[width=.9\linewidth]{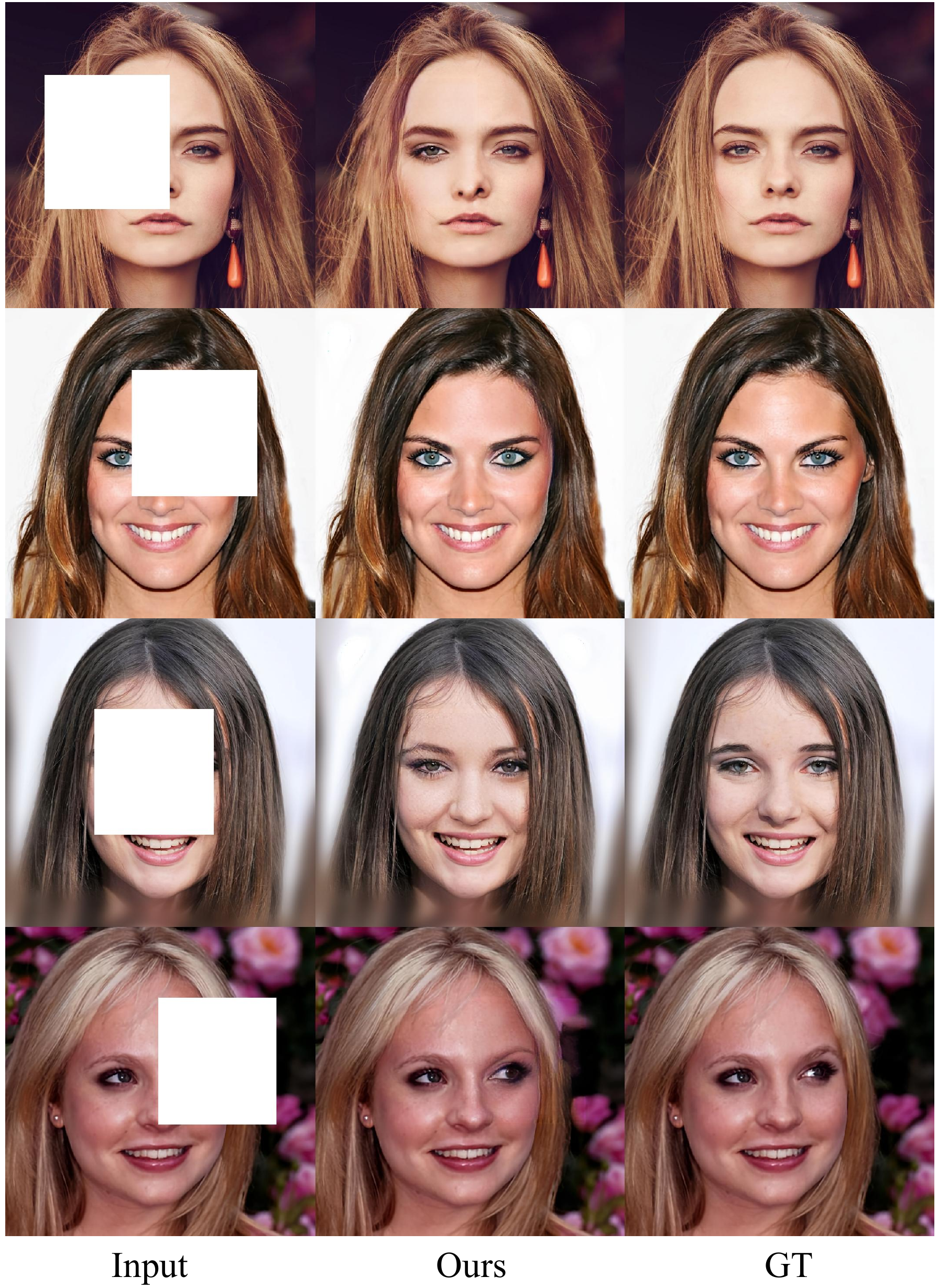}
\end{center}
   \caption{Results on high-resolution images of CelebA-HQ ($1024 \times 1024$).}
\label{fig:10242}
\end{figure*}

\begin{figure*}
\begin{center}
%\fbox{\rule{0pt}{2in} \rule{.7\linewidth}{0pt}}
\includegraphics[width=.9\linewidth]{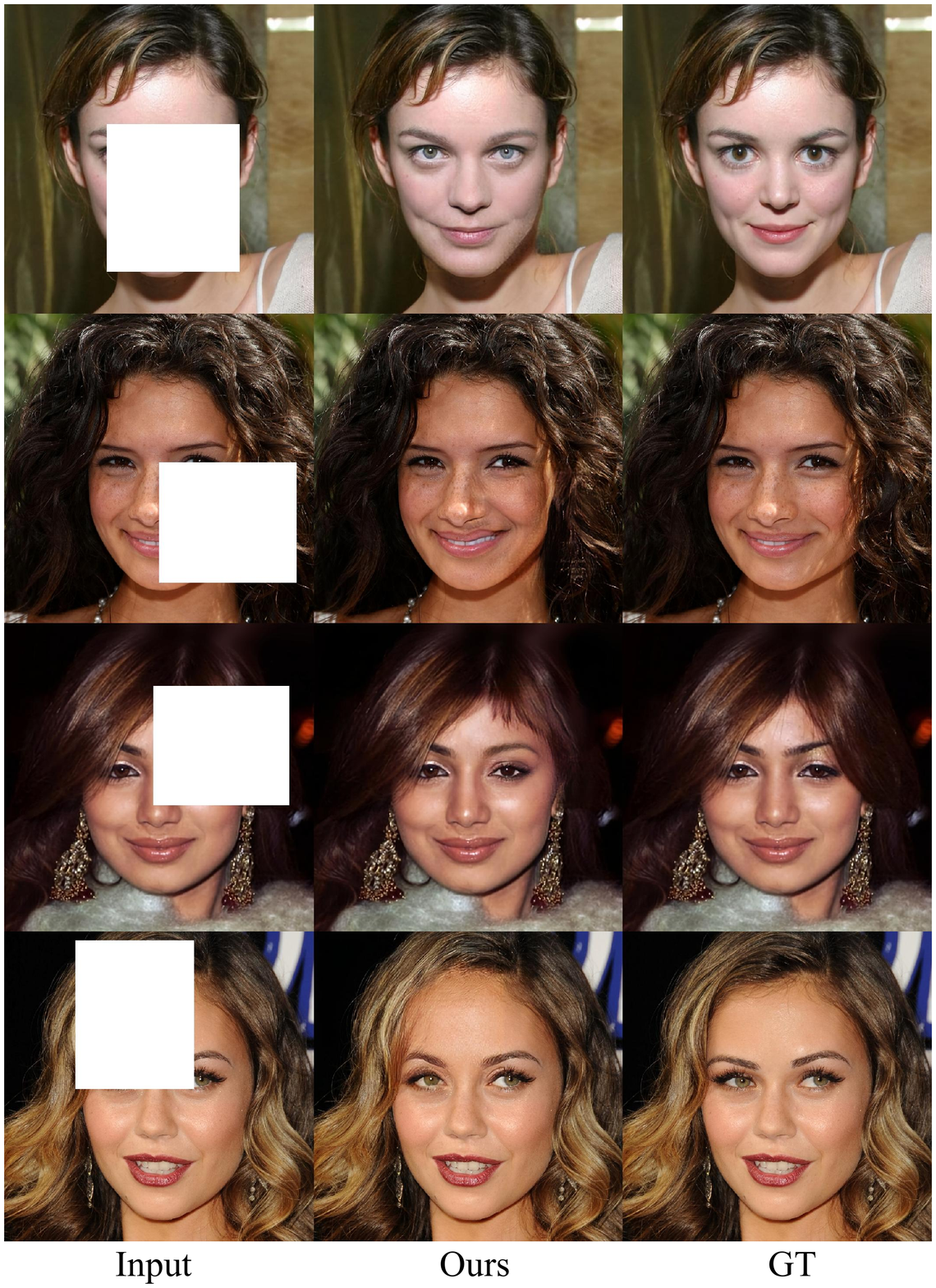}
\end{center}
   \caption{Results on high-resolution images of CelebA-HQ ($1024 \times 1024$).}
\label{fig:10243}
\end{figure*}

In this experiment, we show the effectiveness of our approach on high resolution face images ($1024 \times 1024$).
The experimental settings are consistent with those on low resolution images ($256 \times 256$).
The results are illustrated in Figure~\ref{fig:10241}, Figure~\ref{fig:10242}, Figure~\ref{fig:10243}. It is shown that the recovered images by our approach contain rich facial textures. The facial textures are also highly consistent with the ground-truth images. Therefore, the ability of our approach to generate high-fidelity facial images is justified.
\section{Network Architecture}
\label{sec5}
We show the architecture details of the generator and discriminators in our model in Table~\ref{table:gen} and Table~\ref{table:dis} respectively.  For high resolution images ($1024 \times 1024$), the architecture of our model is adjusted slightly, as shown in Table~\ref{table:gen1024} and Table~\ref{table:dis1024}.
\clearpage

\clearpage

\clearpage

\clearpage

\clearpage

\clearpage

\clearpage

\clearpage

\begin{table*}
\begin{center}
\begin{tabular}{|l|l|}
\hline
Layer 1 & Conv(7, 7, 64), stride=2; ReLU \\
\hline
Layer 2 & Conv(5, 5, 128), stride=2; BN; ReLU \\
\hline
Layer 3 & Conv(3, 3, 256), stride=2; BN; ReLU \\
\hline
Layer 4 & Conv(3, 3, 512), stride=2; BN; ReLU \\
\hline
Layer 5 & Conv(3, 3, 512), stride=2; BN; ReLU \\
\hline
Layer 6 & Conv(3, 3, 512), stride=2; BN; ReLU \\
\hline
Layer 7 & Conv(3, 3, 512), stride=2; BN; ReLU \\
\hline
Layer 8 & Conv(3, 3, 512), stride=2; BN; ReLU \\
\hline
Layer 9 & \makecell[l]{Upsample(factor = 2); Concat(w/ Layer 7); \\ Conv(3, 3, 512), stride=1;  BN; \\LReLU(slope = 0.2);} \\
\hline
Layer 10 & \makecell[l]{Upsample(factor = 2); Concat(w/ Layer 6); \\ Conv(3, 3, 512), stride=1;  BN; \\LReLU(slope = 0.2); }\\
\hline
Layer 11 &\makecell[l]{ Upsample(factor = 2); Concat(w/ Layer 5); \\ Conv(3, 3, 512), stride=1;  BN; \\LReLU(slope = 0.2);} \\
\hline
Layer 12 & \makecell[l]{Upsample(factor = 2); Concat(w/ Layer 4); \\ Conv(3, 3, 512), stride=1;  BN;\\ LReLU(slope = 0.2); \\ Dual Spatial Attention Module(DSA);} \\
\hline
Layer 13 & \makecell[l]{Upsample(factor = 2); Concat(w/ Layer 3); \\ Conv(3, 3, 256), stride=1;  BN;\\ LReLU(slope = 0.2); \\ Dual Spatial Attention Module(DSA);} \\
\hline
Layer 14 & \makecell[l]{Upsample(factor = 2); Concat(w/ Layer 2); \\ Conv(3, 3, 128), stride=1;  BN;\\ LReLU(slope = 0.2); \\ Dual Spatial Attention Module(DSA);} \\
\hline
Layer 15 & \makecell[l]{Upsample(factor = 2); Concat(w/ Layer 1); \\ Conv(3, 3, 64), stride=1;  BN;\\ LReLU(slope = 0.2); } \\
\hline
Layer 16 & \makecell[l]{Upsample(factor = 2); Concat(w/ Input); \\ Conv(3, 3, 3), stride=1; Sigmoid} \\
\hline
\end{tabular}
\end{center}
\caption{The architecture of the generator. BN denotes batch normalization and LReLU denotes leaky ReLU. We adopt a very similar U-Net structure as used in~\cite{liu2018image} for the generator. The difference lies in two aspects: (1) we adopt conventional convolution rather than partial convolution; (2) we equip U-net with the Dual Spatial Attention (DSA) module.}
\label{table:gen}
\end{table*}

\begin{table*}
\begin{center}
\begin{tabular}{|l|l|}
\hline
Layer 1 & Conv(4, 4, $C$), stride=2; LReLU(slope = 0.2); \\
\hline
Layer 2 & Conv(4, 4, 2 $\times$ $C$), stride=2; LReLU(slope = 0.2); \\
\hline
Layer 3 & Conv(4, 4, 4 $\times$ $C$), stride=2; LReLU(slope = 0.2); \\
\hline
Layer 4 & Conv(4, 4, 8 $\times$ $C$), stride=1; LReLU(slope = 0.2); \\
\hline
Layer 5 & Conv(4, 4, 1), stride=1 \\
\hline
\end{tabular}
\end{center}
\caption{The architecture of discriminators. $C$ denotes the number of channels of the convolutional layers. For the local subdivision discriminator and the four organ discriminators imposed on facial components, $C$ equals to 32. For the global and local discriminators, $C$ equals to 64 and 48, respectively.}
\label{table:dis}
\end{table*}
\clearpage

\begin{table*}
\begin{center}
\begin{tabular}{|l|l|}
\hline
Layer 1 & Conv(7, 7, 64), stride=2; ReLU \\
\hline
Layer 2 & Conv(5, 5, 128), stride=2; BN; ReLU \\
\hline
Layer 3 & Conv(3, 3, 256), stride=2; BN; ReLU \\
\hline
Layer 4 & Conv(3, 3, 512), stride=2; BN; ReLU \\
\hline
Layer 5 & Conv(3, 3, 512), stride=2; BN; ReLU \\
\hline
Layer 6 & Conv(3, 3, 512), stride=2; BN; ReLU \\
\hline
Layer 7 & Conv(3, 3, 512), stride=2; BN; ReLU \\
\hline
Layer 8 & Conv(3, 3, 512), stride=2; BN; ReLU \\
\hline
Layer 9 & Conv(3, 3, 512), stride=2; BN; ReLU \\
\hline
Layer 10 & Conv(3, 3, 512), stride=2; BN; ReLU \\
\hline
Layer 11 & \makecell[l]{Upsample(factor = 2); Concat(w/ Layer 9); \\ Conv(3, 3, 512), stride=1;  BN; LReLU(slope = 0.2);} \\
\hline
Layer 12 & \makecell[l]{Upsample(factor = 2); Concat(w/ Layer 8); \\ Conv(3, 3, 512), stride=1;  BN; LReLU(slope = 0.2); } \\
\hline
Layer 13 & \makecell[l]{Upsample(factor = 2); Concat(w/ Layer 7); \\ Conv(3, 3, 512), stride=1;  BN; LReLU(slope = 0.2); } \\
\hline
Layer 14 & \makecell[l]{Upsample(factor = 2); Concat(w/ Layer 6); \\ Conv(3, 3, 512), stride=1;  BN; LReLU(slope = 0.2); \\ Dual Spatial Attention Module(DSA); }\\
\hline
Layer 15 &\makecell[l]{ Upsample(factor = 2); Concat(w/ Layer 5); \\ Conv(3, 3, 512), stride=1;  BN; LReLU(slope = 0.2); \\ Dual Spatial Attention Module(DSA);} \\
\hline
Layer 16 & \makecell[l]{Upsample(factor = 2); Concat(w/ Layer 4); \\ Conv(3, 3, 512), stride=1;  BN; LReLU(slope = 0.2); \\ Dual Spatial Attention Module(DSA);} \\
\hline
Layer 17 & \makecell[l]{Upsample(factor = 2); Concat(w/ Layer 3); \\ Conv(3, 3, 256), stride=1;  BN; LReLU(slope = 0.2);} \\
\hline
Layer 18 & \makecell[l]{Upsample(factor = 2); Concat(w/ Layer 2); \\ Conv(3, 3, 128), stride=1;  BN; LReLU(slope = 0.2);} \\
\hline
Layer 19 & \makecell[l]{Upsample(factor = 2); Concat(w/ Layer 1); \\ Conv(3, 3, 64), stride=1;  BN; LReLU(slope = 0.2); } \\
\hline
Layer 20 & \makecell[l]{Upsample(factor = 2); Concat(w/ Input); \\ Conv(3, 3, 3), stride=1; Sigmoid} \\
\hline
\end{tabular}
\end{center}
\caption{The architecture of the generator for input images of $1024 \times 1024$. To accommodate the high resolution, we add two convolutional layers for both the encoder and decoder of the generator in Table~\ref{table:gen}.}
\label{table:gen1024}
\end{table*}

\begin{table*}
\begin{center}
\begin{tabular}{|l|l|}
\hline
Layer 1 & Conv(4, 4, $C$), stride=2; LReLU(slope = 0.2); \\
\hline
Layer 2 & Conv(4, 4, 2 $\times$ $C$), stride=2; LReLU(slope = 0.2); \\
\hline
Layer 3 & Conv(4, 4, 4 $\times$ $C$), stride=2; LReLU(slope = 0.2); \\
\hline
Layer 4 & Conv(4, 4, 8 $\times$ $C$), stride=2; LReLU(slope = 0.2); \\
\hline
Layer 5 & Conv(4, 4, 8 $\times$ $C$), stride=2; LReLU(slope = 0.2); \\
\hline
Layer 6 & Conv(4, 4, 8 $\times$ $C$), stride=1; LReLU(slope = 0.2); \\
\hline
Layer 7 & Conv(4, 4, 1), stride=1 \\
\hline
\end{tabular}
\end{center}
\caption{The architecture of discriminators for input images of $1024 \times 1024$. To accommodate the high resolution, we add two convolutional layers for discriminators in Table~\ref{table:dis}.}
\label{table:dis1024}
\end{table*}
\clearpage

\end{document}